\documentclass[lettersize,journal]{IEEEtran}
\usepackage{amsmath,amsfonts}
\usepackage{algorithmic}
\usepackage{algorithm}
\usepackage{array}
\usepackage[caption=false,font=normalsize,labelfont=sf,textfont=sf]{subfig}
\usepackage{textcomp}
\usepackage{stfloats}
\usepackage{url}
\usepackage{verbatim}
\usepackage{graphicx}
\usepackage{cite}
\usepackage{enumerate}
\usepackage{amsmath}
\usepackage{color}
\begin{document}

\title{MCMS: Multi-Category Information and Multi-Scale Stripe Attention for Blind Motion Deblurring}

\author{Nianzu~Qiao,~Lamei~Di,~and~Changyin~Sun}

\markboth{Journal of \LaTeX\ Class Files,~Vol.~14, No.~8, August~2022}%
{Shell \MakeLowercase{\textit{et al.}}: A Sample Article Using IEEEtran.cls for IEEE Journals}

\IEEEpubid{0000--0000/00\$00.00~\copyright~2022 IEEE}

\maketitle

\begin{abstract}
Deep learning-based motion deblurring techniques have advanced significantly in recent years. This class of techniques, however, does not carefully examine the inherent flaws in blurry images. For instance, low edge and structural information are traits of blurry images. The high-frequency component of blurry images is edge information, and the low-frequency component is structure information. A blind motion deblurring network (MCMS) based on multi-category information and multi-scale stripe attention mechanism is proposed. Given the respective characteristics of the high-frequency and low-frequency components, a three-stage encoder-decoder model is designed. Specifically, the first stage focuses on extracting the features of the high-frequency component, the second stage concentrates on extracting the features of the low-frequency component, and the third stage integrates the extracted low-frequency component features, the extracted high-frequency component features, and the original blurred image in order to recover the final clear image. As a result, the model effectively improves motion deblurring by fusing the edge information of the high-frequency component and the structural information of the low-frequency component. In addition, a grouped feature fusion technique is developed so as to achieve richer, more three-dimensional and comprehensive utilization of various types of features at a deep level. Next, a multi-scale stripe attention mechanism (MSSA) is designed, which effectively combines the anisotropy and multi-scale information of the image, a move that significantly enhances the capability of the deep model in feature representation. Large-scale comparative studies on various datasets show that the strategy in this paper works better than the recently published measures. 
\end{abstract}


\section{Introduction}
\IEEEPARstart{T}{he} relative motion between the object and the camera is the fundamental factor in the formation of motion blur images. For instance, the camera shakes or moves while the object remains stationary. Alternatively, the object may be moving irregularly or regularly while the camera remains still. Additionally, some practical applications are incompatible with the motion blur image that contains pixel displacement. For example, autonomous driving, unmanned aircraft, unmanned ships, and intelligent surveillance systems. Additionally, a number of advanced semantic activities are adversely affected by blurry images. For instance, image classification, semantic segmentation, information dissemination \cite{ref6}, and object recognition, etc. Therefore, the deblurring technique of motion blur images has been a research highlight in the field of computer vision.
\IEEEpubidadjcol

Image motion deblurring has been approached in a number of different ways. These tactics have been organized into two main groups: deep learning-based procedures and conventional techniques depend on a priori. Below is a detailed list of both sorts of approaches' development timelines.

Traditional methods based on prior. Pan et al. \cite{ref10} suggestd a blind motion deblurring solution derived from dark channel prior, which exploits dark channel sparsity prior to repair blurry images. An L0 sparse expression that can successfully eliminate motion blur was created by Xu et al. \cite{ref11}. A combined channel prior created by Yan et al. \cite{ref12} and formed from dark and bright channels prior can successfully eliminate motion blur. Chen et al. \cite{ref13} designed a motion deblurring arrangement derived from the local gradient maximum prior. Pan et al. \cite{ref14} designed a text image deblurring technique derived from L0 regularization prior of intensity and gradient. Dong et al. \cite{ref15} designed a approach to handle motion blur image outliers. Bahat et al. \cite{ref16} utilized quadratic blurring to thoroughly evaluate the content information of blurry images and the measure can effectively remove motion blur. Sheng et al. \cite{ref17} derived from depth map to effectively estimate the blur kernel for the purpose of blurry image restoration.

Although the above traditional methods based on a priori have achieved certain deblurring effects, this class of ways also have several disadvantages: 1) this type of plans necessitates multiple a priori information, necessitating immensely challenging mathematical formulae for derivation. 2) These procedures have severe limits since they demand high-quality a priori knowledge. 3) Non-uniformly blurry images cannot be handled by this kind of strategy. Because the blur kernel of non-uniformly blurry images cannot be accurately estimated using conventional approaches.

Deep learning-based approach. Currently, deep learning has achieved remarkable success in the field of image enhancement. Examples include dehazing \cite{ref18}, water removal \cite{ref19}, and super-resolution reconstruction \cite{ref20,ref21}. In the meantime, image deblurring has benefited from some advancements in deep learning. For the first time, Chakrabarti et al. \cite{ref22} exploited pre-trained deep neural networks to estimate sharp images. Nah et al. \cite{ref23} designed a multi-scale neural network to remove motion blur end-to-end. Tao et al. \cite{ref24} designed a recursive network by combining scale structures. Despite having fewer parameters, this network requires more time to train as a result of recursion's convergence property. Gao et al. \cite{ref25} designed a selective sharing technique and incorporated skip connection to the internal submodule part. Zhang et al. \cite{ref26} advised a multi-scale multi-patch network. This network's deblurring ability has slightly increased compared to the multi-scale network \cite{ref23}. Park et al. \cite{ref27} suggested an alternative multi-scale multi-temporal deblurring plan. Additionally, the technique includes recursive elements. Zamir et al. \cite{ref28} carried out an improved design based on \cite{ref26}, specifically by laterally shifting the network structure in \cite{ref26}. Esmaeilzehi et al. \cite{ref29} designed a lightweight residual network based on upsampling and deblurring modules. Ji et al. \cite{ref30} suggested a single encoder-dual decoder network configuration to eliminate motion blur.

Although the currently suggested deep learning-based solutions have achieved splendid outcomes, the aforementioned plans still have certain shortcomings. Particularly, the suggested deep learning-based methods emphasize the network structure more than the properties of the motion blur images, which are investigated less thoroughly. Liu et al. \cite{ref31} incorporated high-frequency (HF) information into the deblurring network, and the plan achieved sensational results. However, this way simply considered the high-frequency information, ignoring the significance of low-frequency (LF) information for motion deblurring. Although the sharpness of the image is immediately impacted by the quality of the HF information, the LF information constitutes the bulk of the image's structure. As a result, the quality of the LF information is tightly tied to the sharpness of the image.

A motion deblurring network that integrates HF information with LF information of images is suggested as a result of the aforementioned findings.

\section{Related Works}
\subsection{Physical Model of Motion Blur Images}
From the literature \cite{ref32}, the physical model of the motion blur images is shown in Eq. 1.

\begin{equation}
	\label{deqn_ex1}
	b=I\otimes k + n.
\end{equation}

Where $b$ is the blurry image, $I$ is the sharp image, $k$ is the blur kernel, and $n$ is the noise, $\otimes$ denotes the convolution operation. As well known as blind motion blur, this work deals with the situation when the blur kernel $k$ is unknown. Therefore, to remove the blind motion blur, it is necessary to recover both the sharp image $I$ and the blur kernel $k$. The conventional algorithm recovers the sharp image $I$ by estimating the blur kernel $k$, while the deep learning algorithm is end-to-end to recover the sharp image $I$.

\subsection{HF Component and LF Component of the Image}
It is straightforward to determine from the frequency of the sound: HF implies high-pitch, such as birdsong or violins. On the other hand, LF is low-pitch, such as a low voice or a bass drum. The frequency of the sound is the rate at which the sound wave oscillates. Where oscillation is usually measured in cycles per second (Hz). This leads to the conclusion that high-pitch is produced by HF wave and low-pitch is produced by LF wave.

Similar comparisons can be made between sound and vision. The rate of change of the pixel is what is known as the frequency in an image. The image undergoes a series of changes in the spatial dimension. Edge contours and texture information are described in HF images, which are images with rapid regional changes. Additionally, LF images are images with gradual regional shifts that convey the primary information in the image (structure and content).

Currently, there are numerous methods for dividing an image into HF and LF components. For illustrate, the discrete cosine transform \cite{ref33}, wavelet transform \cite{ref34}, and Framelet \cite{ref35}. In this study, the discrete cosine transform is applied to segregate the image's HF and LF components.

\section{The recommended approach}
In this section, an image motion deblurring processing technique called MCMS is designed as shown in Fig. 1. MCMS is derived from an encoder-decoder structure, where the encoder part contains three dimensions of information processing and the decoder has three dimensions of information reconstruction.

\begin{figure*}[t]
	\centering
	\includegraphics[width=6in]{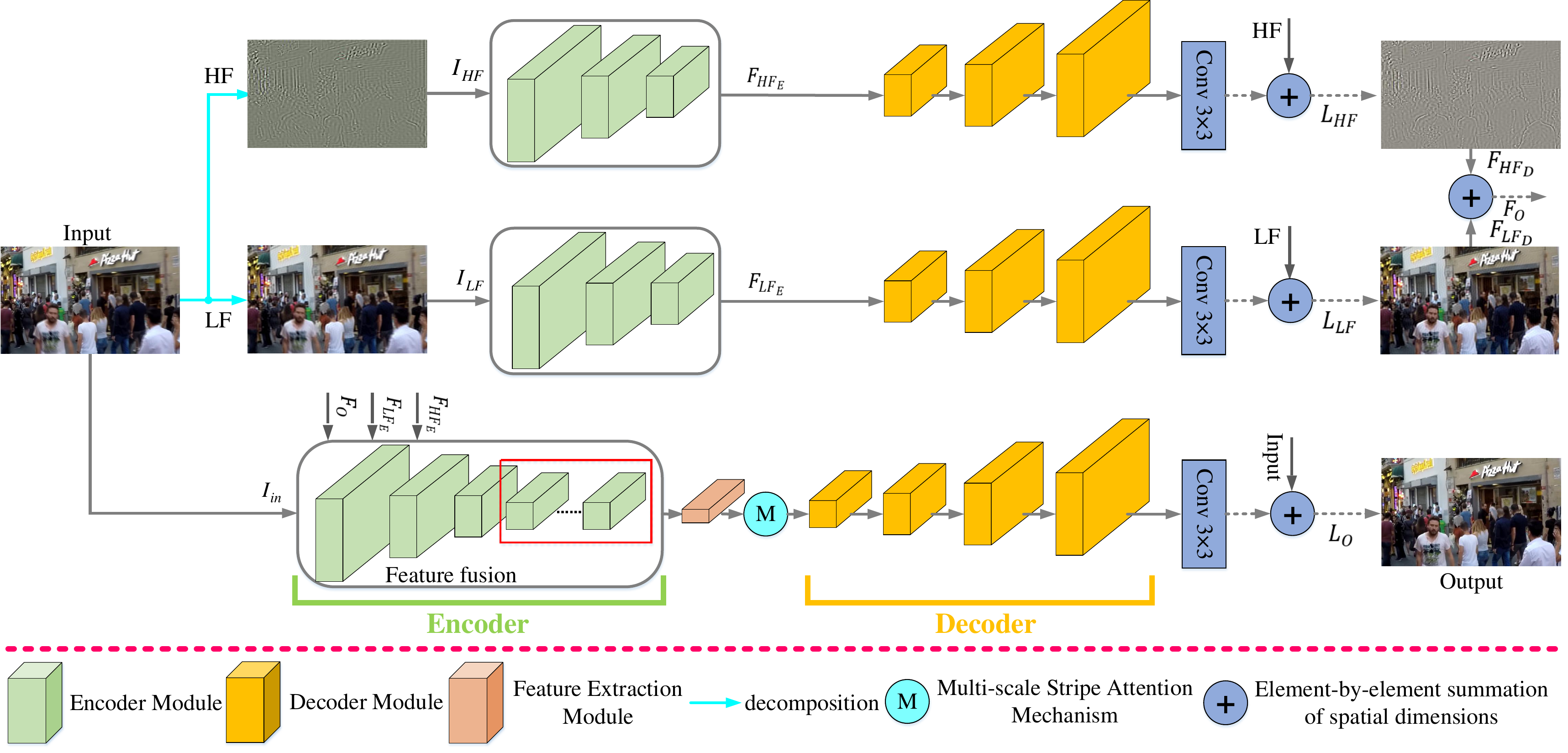}
	\caption{The overall composition of MCMS.}
	\label{fig1}
\end{figure*}

The High-Frequency component (HF), Low-Frequency component (LF) and the original blurred image together form these three dimensions. In this paper, the design of the encoding and decoding modules is borrowed from the Block module in NAFNet [113]. Specifically, the encoder and decoder of the high-frequency component and low-frequency component branches each employ three Block modules. And the red box part of the encoder of the third branch, which is responsible for fusing the three kinds of information, specifically employs 28 Block modules. In order to fully exploit the value of various types of information, this paper proposes an innovative feature fusion strategy, which is summarized below.

\subsection{MCMS Construction Details}
Based on Retinex theory \cite{ref191}, it can be known that the HF component of an image reflects the edge and texture information of the image, while the LF component represents the content and structure information of the image, as shown in Fig. 2. Inspired by this, for the HF component and LF component of motion blurred images, we adopt a staged processing strategy.

\begin{figure}[h]
	\centering
	\includegraphics[width=3.3in]{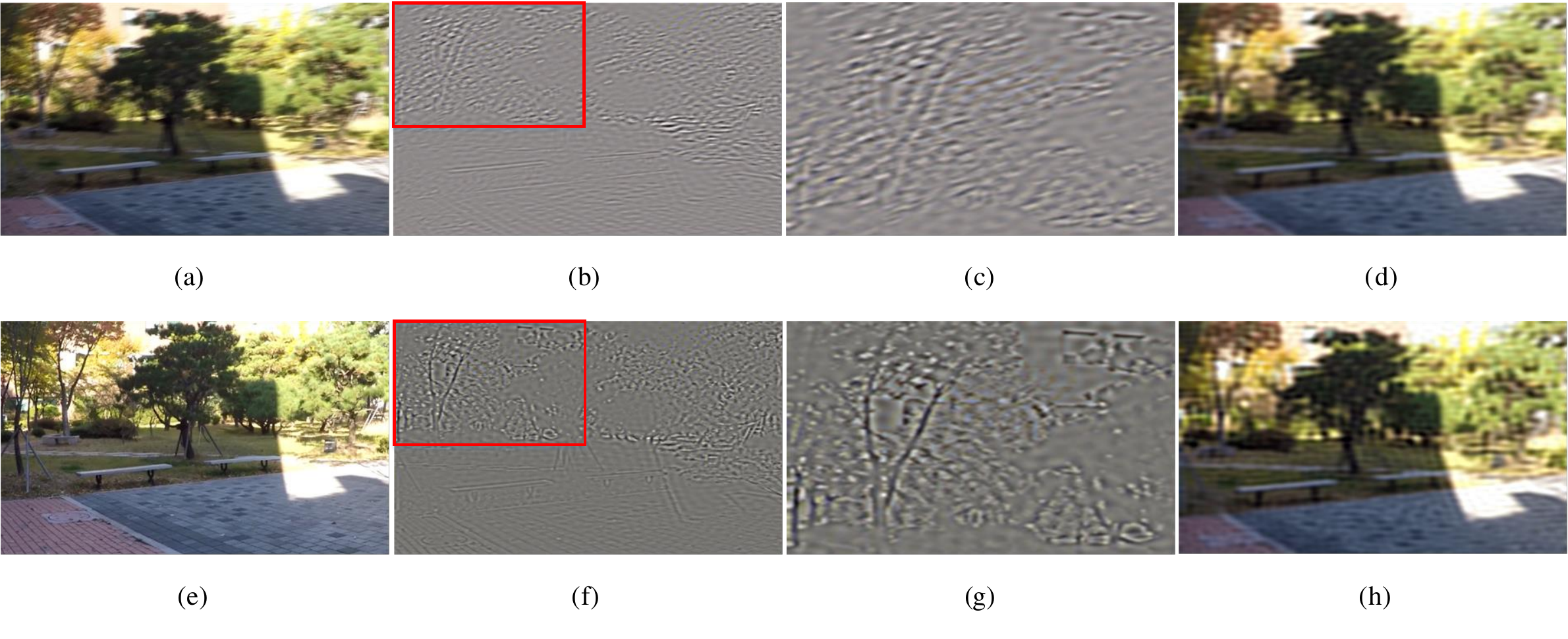}
	\caption{HF component and LF component of the blurry and sharp images. (a) Blurry image.(b) Blurry HF component. (c) Locally enlarged version of the blurry HF component. (d) Blurry LF component. (e) Sharp image.(f) Sharp HF component. (g) Locally enlarged version of the sharp HF component (h) Sharp LF component.}
	\label{fig2}
\end{figure}

The LF component represents the structural and contextual information of an image. It embodies the characteristics of large, smooth regions in the image with relatively few changes and details, and plays an important role in processing the basic elements of the image. Combining the characteristics of the LF component and the structural self-similarity a priori knowledge of the image \cite{ref192}, MCMS focuses on the extraction of the structural information of the LF component, so that it can recover the structural information of the image more accurately.The HF component mainly reflects the detail and change information of the image. It reflects the properties of localized regions in the image, such as texture, edges, etc. The HF component usually contains more changes and details and plays an important role in processing the detailed features of the image. Inspired by the own properties of HF components, MCMS pays more attention to the extraction of local detail information of HF components, including edge and texture information, so that it can recover the edge and texture information of the image more accurately.

The MCMS's unique design principles include:

First, the encoder and decoder structures in the first and second stages are used for the processing of the HF component and the LF component. Subsequently, the original blurred image is processed by the encoder and decoder structures in the third stage. In addition, the three types of features are also fused in a multidimensional and all-encompassing way to maximize the advantages of each type of feature.

\subsection{Grouped Feature Fusion Module}

In order to fully exploit various types of feature information to enhance network performance, a grouped feature fusion strategy is designed as shown in Fig. 3. Eq. 2 shows the specific mathematical derivation process.

\begin{figure}[h]
	\centering
	\includegraphics[width=2.3in]{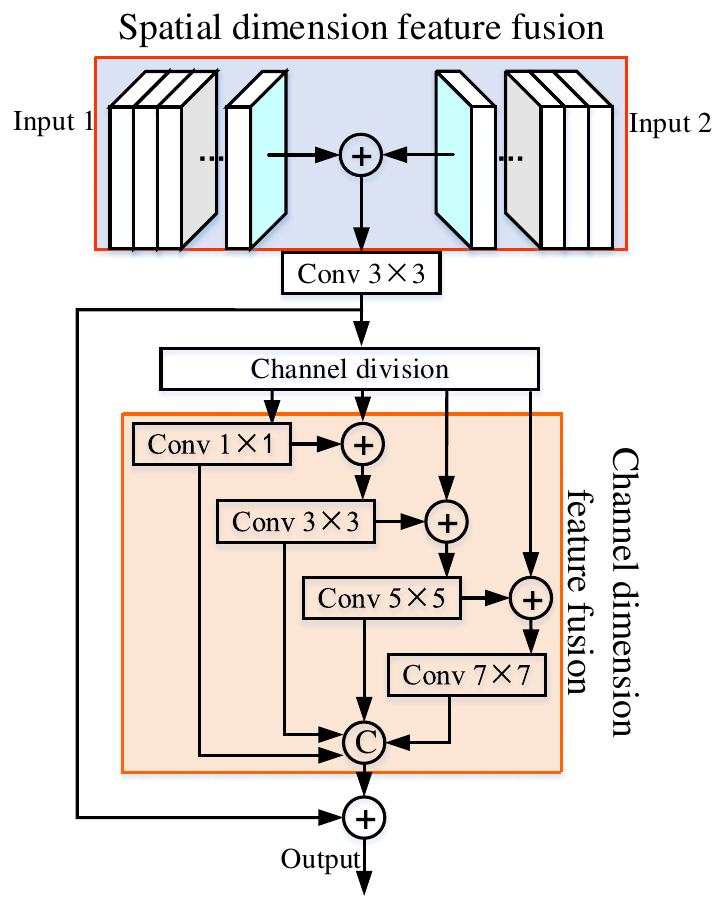}
	\caption{Example of grouped feature fusion.}
	\label{fig3}
\end{figure}

\begin{equation}
	\begin{aligned}
			& [conv1\times 1\left( {{C}_{1}} \right), \\ 
			& conv3\times 3\left( conv1\times 1\left( {{C}_{1}} \right)+{{C}_{2}} \right), \\ 
			& conv5\times 5\left( conv3\times 3\left( conv1\times 1\left( {{C}_{1}} \right)+{{C}_{2}} \right),{{C}_{3}} \right), \\ 
			& conv7\times 7( conv5\times 5( conv3\times 3\left( conv1\times 1\left( {{C}_{1}} \right)+{{C}_{2}} \right), \\
			&{{C}_{3}}),{{C}_{4}} )] \\ 
		& +conv3\times 3\left( {{I}_{1}}+{{I}_{2}} \right) \\ 
		& {{C}_{1}},{{C}_{2}},{{C}_{3}},{{C}_{4}}=chunk\left( conv3\times 3\left( {{I}_{1}}+{{I}_{2}} \right) \right). 
	\end{aligned}
\end{equation}

where $\left[ \cdot ,\cdot ,\cdot ,\cdot  \right]$ represents the Concat operation. $conv1\times 1$ represents the $1\times 1$ convolution operation, $conv3\times 3$ represents the $3\times 3$ convolution operation, $conv5\times 5$ represents the $5\times 5$ convolution operation, $conv7\times 7$ represents the $7\times 7$ convolution operation. $chunk$ represents the channel equal division operation.

The method fuses the information of the input feature maps in the spatial dimension and the channel dimension. In the spatial dimension, the two input feature maps are fused by element-by-element summation. This approach enables the new feature map to fully integrate the information of the two input feature maps, thus showing richer characteristics in the spatial dimension. In the channel dimension, the feature map is divided into four sub-channels for convolutional computation at four scales. This operation effectively extracts the multi-scale information in the feature map, enabling the model to capture the details and structure of the image at different scales more comprehensively. It also enables the model to understand the features of the image at different levels. Overall, grouped feature fusion effectively improves the feature extraction capability of MCMS by finely extracting features in both the channel dimension and the spatial dimension.

As can be seen from Fig. 1, feature fusion mainly occurs in the third stage: feature fusion of the HF component, the LF component and the original blurred image feature map.

Fig. 4 illustrates the feature fusion process. Feature fusion helps to obtain richer, more three-dimensional and more comprehensive feature information, which further improves the performance of the whole network. Eq. 3 shows the specific mathematical derivation process.

\begin{figure}[bht]
	\centering
	\includegraphics[width=2.5in]{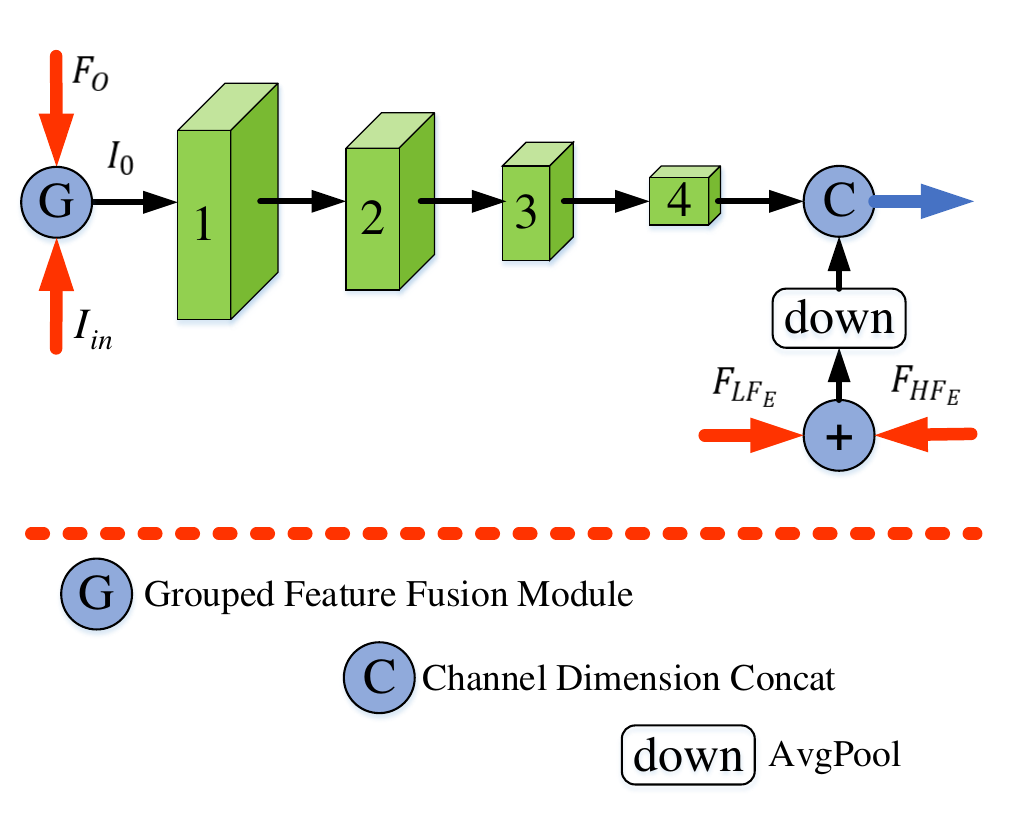}
	\caption{Feature fusion process.}
	\label{fig4}
\end{figure}

\begin{equation}
	\begin{aligned}
	& {{F}_{E}}=\left[ {{E}_{3}}\left( {{E}_{2}}\left( {{E}_{1}}\left( {{I}_{0}} \right) \right) \right),d\left( {{F}_{L{{F}_{E}}}}+{{F}_{H{{F}_{E}}}} \right) \right], \\ 
	& {{I}_{0}}=G\left( {{I}_{in}},{{F}_{O}} \right), \\ 
	& {{F}_{O}}={{F}_{H{{F}_{D}}}}\oplus {{F}_{L{{F}_{D}}}}.
	\end{aligned}
\end{equation}

where $\left[ \cdot ,\cdot  \right]$ represents the Concat operation. ${{I}_{0}}$ represents the feature fusion result of LF component, HF component and original blurred image. ${{I}_{in}}$ represents the original blurred image. ${{F}_{H{{F}_{E}}}}$ represents the encoder output features of the HF component. ${{F}_{L{{F}_{E}}}}$ represents the encoder output features of the LF component. $d$ represents the average pooling operation. $G$ represents grouped feature fusion. ${{F}_{O}}$ denotes the sum of ${{F}_{H{{F}_{D}}}}$ (decoder output features of HF component) and ${{F}_{L{{F}_{D}}}}$ (decoder output features of LF component). $\oplus $ denotes the element-by-element summation operation in spatial dimension. ${{E}_{1}}$, ${{E}_{2}}$, and ${{E}_{3}}$ represent the feature extraction operations of encoder 1, 2, and 3, respectively.

\subsection{Multi-scale Stripe Attention Mechanism}
Li et al. \cite{ref203} proposed a stripe self-attention mechanism based on image anisotropy property, which is an effective vertical and horizontal stripe self-attention. Therefore, we further evolve this mechanism into a stripe attention mechanism applicable to CNN models. Although this mechanism can improve the accuracy of the model to a certain extent, its limitation is that it only considers the single-scale stripe attention and ignores the actual multi-scale information.

Many current studies have demonstrated the advantages of multiscale. Multiscale methods possess the ability to efficiently extract information at different scales and help expand the perceptual range of the extracted information. Based on this background, we design a multiscale striped attention mechanism, which skillfully combines image anisotropy and multiscale information, thus further enhancing the feature representation capability of the depth model.

We innovatively design a multi-scale stripe attention mechanism. The mechanism has the capability of selectively extracting the multi-scale spatial weights of the feature map. Its unique feature is that it eliminates the effects of spatial distance and single scale of similar features, which helps connectivity between similar fuzzy regions. Overall, the multiscale stripe attention mechanism demonstrates three significant advantages. First, it utilizes the anisotropic properties of images to extend the receptive field of the attention mechanism. Second, the mechanism skillfully incorporates the advantages of multi-scale to effectively enhance the generalization performance of the attention mechanism. Finally, it implicitly enhances the weight of edge information, a property that plays an important role in the field of image deblurring. This is because the motion blur of an image is mainly concentrated on the edge information, and that blur has similarity.

\begin{figure*}[htb]
	\centering
	\includegraphics[width=7in]{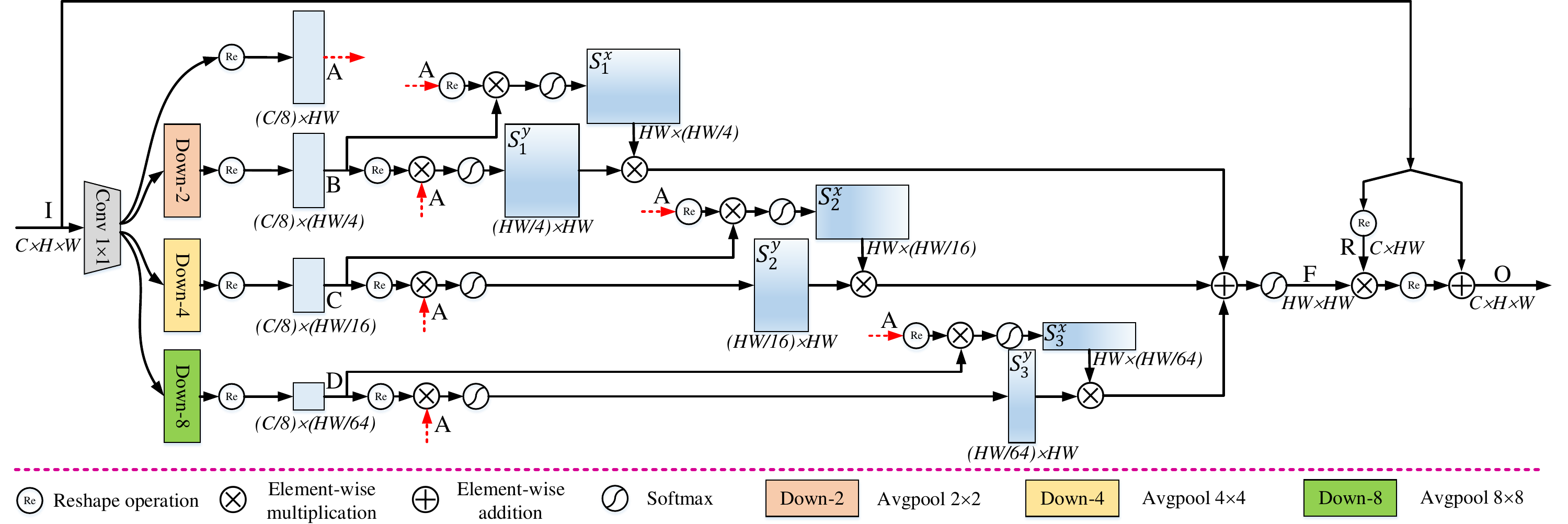}
	\caption{Feature fusion process for the three locations. (a) The feature fusion process for the first position.
		(b) The feature fusion process for the second position. (c) The feature fusion process for the third position.}
	\label{fig5}
\end{figure*}

Fig. 5 clearly shows the operation flow of the whole multi-scale stripe attention mechanism. Firstly, the input feature map $I$ of size $C\times H\times W$ is $Conv1\times 1$ to generate the feature map $\widehat{I}$ of ${C}/{8}\;\times H\times W$. Then, $\widehat{I}$ is directly converted to $\left( {C}/{8}\; \right)\times HW$'s feature map $A$. At the same time, $\widehat{I}$ generates $\left( {C}/{8}\; \right)\times \left( {HW}/{4}\; \right)$'s feature map $B$, $\left( {C}/{8}\; \right)\times \left( {HW}/{16}\; \right)$'s feature map $C$, and $\left( {C}/{8}\; \right)\times \left( {HW}/{64}\; \right)$'s feature map $D$ through Avgpool and Reshape operations at three sizes, $2\times 2$, $4\times 4$, and $8\times 8$, respectively, as shown in the following expressions.

\begin{equation}
\begin{aligned}
	 & A={{R}_{e}}\left( {{C}_{1\times 1}}\left( I \right) \right), \\ 
	& B={{R}_{e}}\left( {{D}_{2\times 2}}\left( {{C}_{1\times 1}}\left( I \right) \right) \right), \\ 
	& C={{R}_{e}}\left( {{D}_{4\times 4}}\left( {{C}_{1\times 1}}\left( I \right) \right) \right), \\ 
	& D={{R}_{e}}\left( {{D}_{8\times 8}}\left( {{C}_{1\times 1}}\left( I \right) \right) \right).
\end{aligned}
\end{equation}

where ${{R}_{e}}$ stands for Reshape operation. ${{D}_{2\times 2}}$, ${{D}_{4\times 4}}$ and ${{D}_{8\times 8}}$ stand for $2\times 2$-scale Avgpool, $4\times 4$-scale Avgpool and $8\times 8$-scale Avgpool respectively. ${{C}_{1\times 1}}$ is $Conv1\times 1$.

Next, $A$ is processed by Reshape and multiplied directly by $B$ and processed by Softmax function to generate the horizontal stripe attention weight matrix $S_{1}^{x}$ of size $HW\times \left( {HW}/{4}\; \right)$. Similarly, $B$ is processed by Reshape and multiplied directly by $A$ and processed by Softmax function to generate the longitudinal stripe attention weight matrix $S_{1}^{y}$ of size $\left( {HW}/{4}\; \right)\times HW$. This analogy generates the horizontal stripe attention weight matrix $S_{2}^{x}$ with size $HW\times \left( {HW}/{16}\; \right)$, the horizontal stripe attention weight matrix $S_{3}^{x}$ with size $HW\times \left( {HW}/{64}\; \right)$, the vertical stripe attention weight matrix $S_{2}^{y}$ with size $\left( {HW}/{16}\; \right)\times HW$, and the vertical stripe attention weight matrix $S_{3}^{y}$ with size $\left( {HW}/{64}\; \right)\times HW$. The specific expression is as follows:

\begin{equation}
	\begin{aligned}
		& S_{1}^{x}=S\left( {{R}_{e}}\left( A \right)\times B \right), \\ 
		& S_{1}^{y}=S\left( {{R}_{e}}\left( B \right)\times A \right), \\ 
		& S_{2}^{x}=S\left( {{R}_{e}}\left( A \right)\times C \right), \\ 
		& S_{2}^{y}=S\left( {{R}_{e}}\left( C \right)\times A \right), \\ 
		& S_{3}^{x}=S\left( {{R}_{e}}\left( A \right)\times D \right), \\ 
		& S_{3}^{y}=S\left( {{R}_{e}}\left( D \right)\times A \right).
	\end{aligned}
\end{equation}

where $S$ denotes the Softmax function.

In the next step, $S_{1}^{x}$ is multiplied by $S_{1}^{y}$, $S_{2}^{x}$ is multiplied by $S_{2}^{y}$, and $S_{3}^{x}$ is multiplied by $S_{3}^{y}$, and the three results obtained are summed. Then, the summed results are processed by Softmax function to generate the $HW\times HW$'s attention weight matrix $F$. Next, $I$ is Reshape processed to get $R$ of $C\times HW$. Then, $R$ and $F$ are multiplied and Reshape processed to generate the attention feature map. Finally, the attention feature map is summed with the residuals $I$ to get the final output $O$. The specific mathematical expression is as follows.

\begin{equation}
	\begin{aligned}
		O={{R}_{e}}\left( R\times S\left( S_{1}^{x}\times S_{1}^{y}+S_{2}^{x}\times S_{2}^{y}+S_{3}^{x}\times S_{3}^{y} \right) \right)+I.
	\end{aligned}
\end{equation}

\subsection{Loss Function}
The loss function in this paper contains three components, $L_{HF}$, which evaluates the quality of the HF
component, $L_{LF}$, which evaluates the quality of the LF component, and $L_O$, which evaluates the quality
of the restored image. Where $L_{HF}$ consists of $L_1$ loss, the following are its specific expressions.

\begin{equation}
	\begin{aligned}
		{{L}_{HF}}={{L}_{1}}=\left\| {{X}_{HF}}-{{Y}_{HF}} \right\|.
	\end{aligned}
\end{equation}

\begin{equation}
	\begin{aligned}
		{{L}_{LF}}={{L}_{1}}=\left\| {{X}_{LF}}-{{Y}_{LF}} \right\|.
	\end{aligned}
\end{equation}

Where $X_{HF}$ is the restored HF component, $Y_{HF}$ is the corresponding HF component of ground
truth, $X_{LF}$ is the restored LF component, $Y_{LF}$ is the corresponding LF component of ground
truth.

$L_O$ consists of $L_1$ loss and MSFR loss \cite{ref38}, the specific expression is shown below.

\begin{equation}
	\begin{aligned}
		L_{O}=&L_1+L_{MSFR}=\parallel X-Y\parallel \\
		&+\gamma\parallel\varphi(X)-\varphi(Y)\parallel.
	\end{aligned}
\end{equation}

Where $X$ is the restored image, $Y$ is the corresponding ground truth, $\gamma = 0.1$ is the weighting
factor. $\varphi$ is the fast Fourier transform (FFT) that transfers image signal to the frequency domain. $L_O$
can evaluate the quality of the recovered image in both the time and frequency domains.

Finally, the final loss function $L_T$ is derived by combining the three types of loss functions, as
shown in Eq. 8.

\begin{equation}
	\begin{aligned}
		L_{T}=L_{HF}+L_{LF}+L_O.
	\end{aligned}
\end{equation}

\begin{figure*}[h]
	\centering
	\includegraphics[width=5.5in]{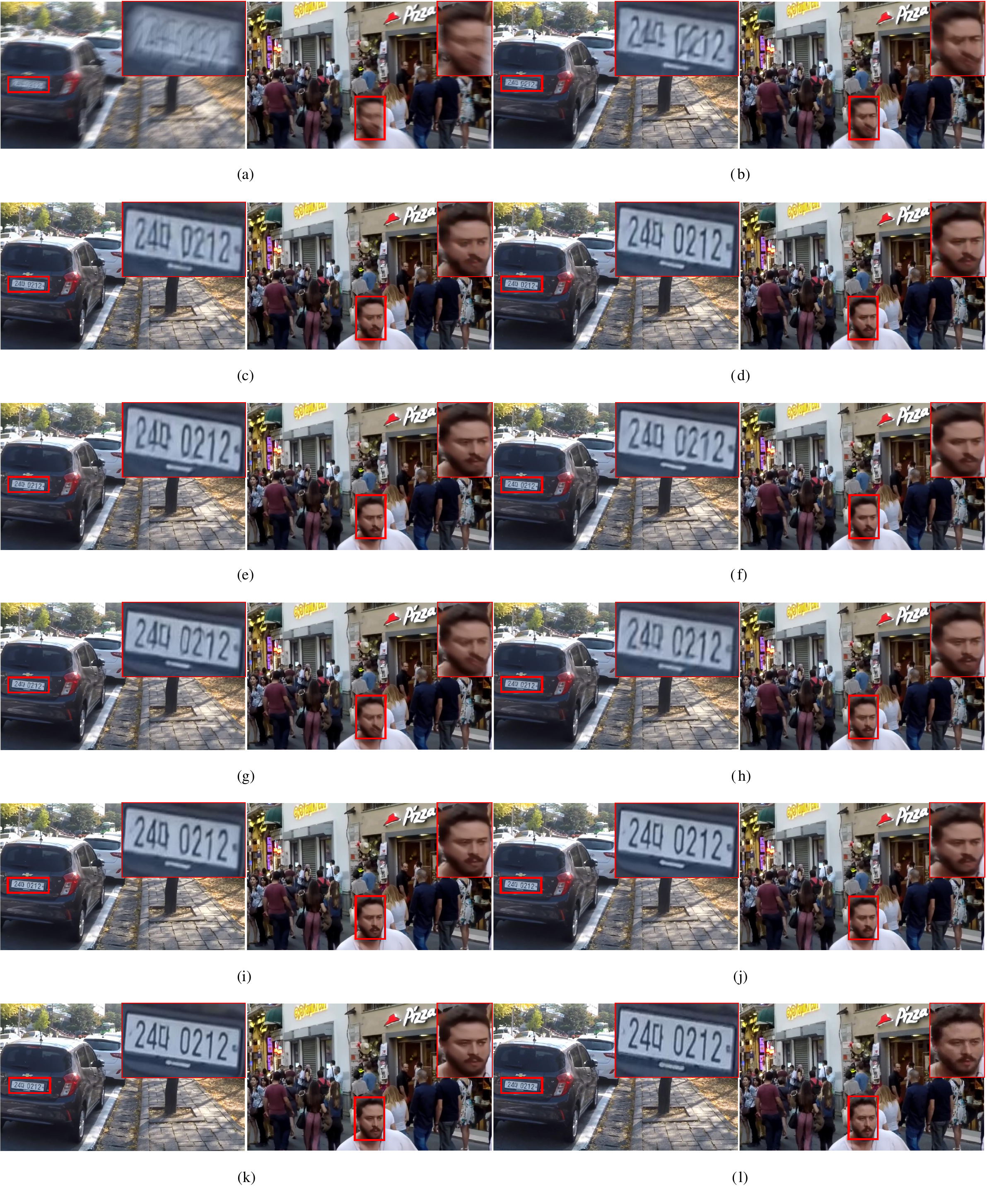}
	\caption{Qualitative results of different ways on the GoPro dataset. (a) Blurry images. (b) DeepDeblur \cite{ref23}. (c) SRN \cite{ref24}. (d) PSS-NSC \cite{ref25}. (e) DMPHN \cite{ref26}. (f) MT-RNN \cite{ref27}. (g) MPR-Net \cite{ref28}. (h) XYDeblur \cite{ref30}. (i) NAFNet \cite{ref113}. (j) MSFS-Net \cite{ref114}. (k) MCMS. (l) Ground-truth.}
	\label{fig6}
\end{figure*}
\begin{figure*}[b]
	\centering
	\includegraphics[width=5in]{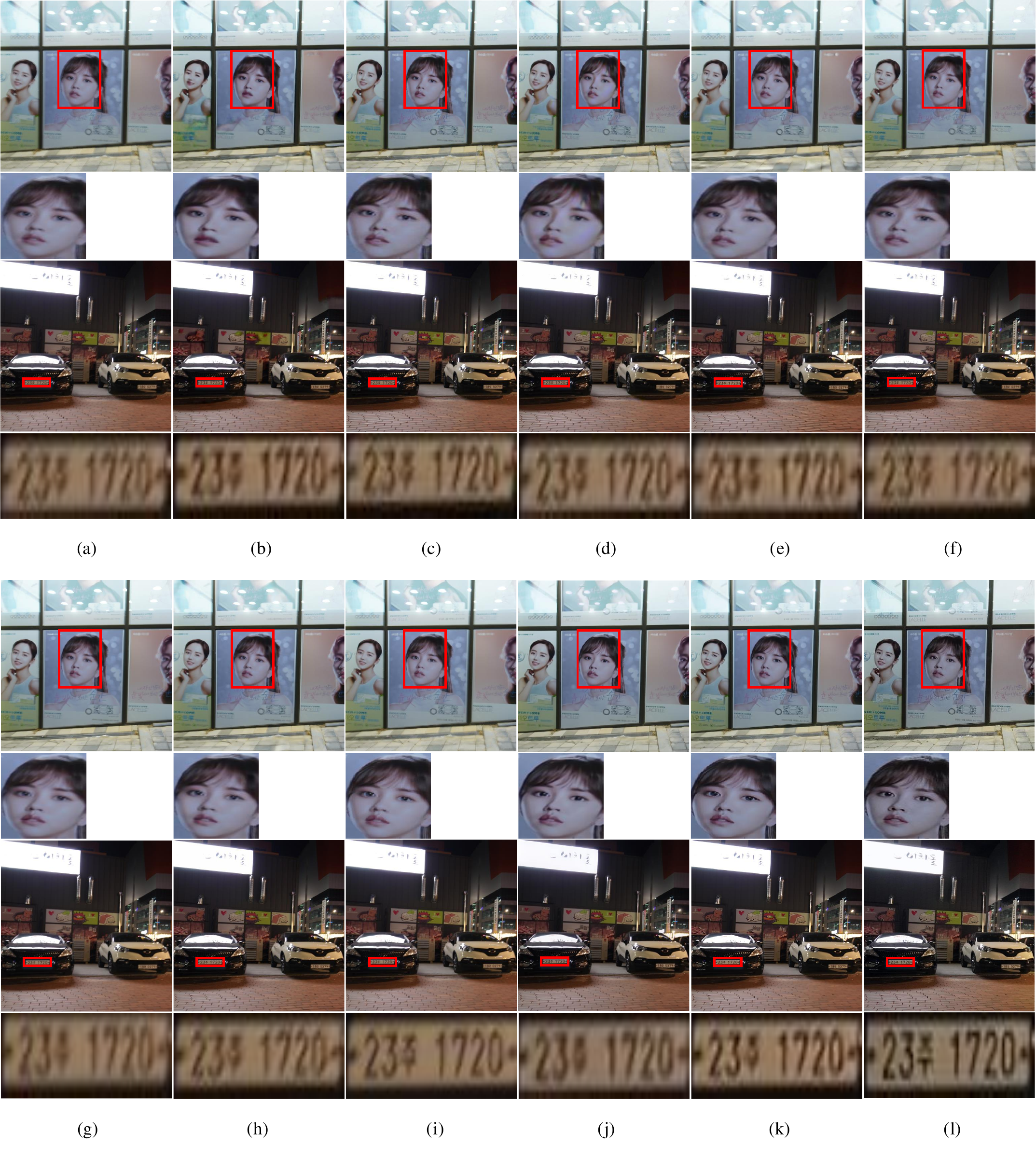}
	\caption{Qualitative results of different plans on the RealBlur dataset. (a) Blurry images. (b) DeepDeblur \cite{ref23}. (c) SRN \cite{ref24}. (d) PSS-NSC \cite{ref25}. (e) DMPHN \cite{ref26}. (f) MT-RNN \cite{ref27}. (g) MPR-Net \cite{ref28}. (h) XYDeblur \cite{ref30}. (i) NAFNet \cite{ref113}. (j) MSFS-Net \cite{ref114}. (k) MCMS. (l) Ground-truth.}
	\label{fig7}
\end{figure*}
\section{Experiments}
In the experimental section, a variety of current state-of-the-art deep learning algorithms are employed for relevant comparative studies, including DeepDeblur \cite{ref23}, SRN \cite{ref24}, PSS-NSC \cite{ref25}, DMPHN \cite{ref26}, MT-RNN \cite{ref27}, MPR-Net \cite{ref28}, XYDeblur \cite{ref30}, NAFNet \cite{ref113} and MSFS-Net \cite{ref114}. Next, the experiments will be described in detail, including qualitative and quantitative comparison tests, ablation analysis on atmospheric image datasets (containing three publicly available datasets) and underwater image datasets (collected and produced by ourselves), in order to comprehensively evaluate the algorithms' performance and effectiveness.

\subsection{Experimental Settings}
The Pytorch framework is utilized to implement MCMS, which is trained on an NVIDIA GeForce RTX 3090 graphics card. The batch size is set to 8 during the training phase, and the learning rate is set to 0.0001. Furthermore, the size of the input images is randomly cropped to $256\times256$ in order to boost training efficiency.

\subsection{Datasets}
The GoPro \cite{ref23} dataset is exploited to train and test our model. It consists of 3214 blurry and sharp image pairs. Where the images size are $1280\times720$. 2103 image pairs from the GoPro dataset are employed to train the model. Meanwhile, assess the generalizability of the model. The trained model on the GoPro dataset is used for testing on the RealBlur \cite{ref41} and RWBI \cite{ref205} dataset. Whereas the RealBlur dataset consists of 4738 blurry and sharp image pairs. The RWBI dataset contains a total of 3112 blurred images. Since there is no corresponding ground truth in the RWBI dataset, only qualitative comparison experiments can be conducted in this dataset.

\subsection{Assessment Indicators}
In this paper, SSIM and PSNR are adopted as the quantitative indexes of the experimental results. Among them, SSIM is a metric to measure structural similarity. It evaluates the image quality in three aspects of brightness, contrast, and structure respectively in a stereoscopic way. The value range of SSIM is [0, 1], and a larger value means better image quality. PSNR mainly measures the magnitude of the error between the recovered image and Ground-truth. Therefore, it is an error-sensitive measure. Its larger value indicates that the image quality is closer to Ground-truth.

\subsection{Performance Comparison}
In this research, we conduct qualitative and quantitative experiments on the GoPro and RealBlur datasets, respectively. Qualitative experiment is conducted on the RWBI dataset. The in-depth experimental findings are displayed below.

\subsubsection{Comparison Experiments on The GoPro Dataset}
\textbf{Qualitative results.} Fig. 6 presents the qualitative evaluation results for the GoPro test dataset. From the figure, it can be seen that compared to other models, MCMS demonstrates significant advantages in processing both the de-blurring effect and the quality of visual perception. The other models have more or less residual motion blur problems in the processing results, which are not completely eliminated. Taking the face region as an example, the performances of various models are evaluated meticulously.

\textbf{Quantitative results.} Table 1 presents the quantitative results of the different plans on the GoPro dataset. It is quite evident that MCMS excels in both evaluation metrics.

\begin{table}[!t]
	\caption{Quantitative results of different methods on the GoPro dataset. Red values indicate the best results, blue values represent sub-optimal performance\label{tab:table1}}
	\centering
	\begin{tabular}{c c c}
		\hline
		Methods & PSNR ($\uparrow$)& SSIM ($\uparrow$)\\
		\hline
		DeepDeblur \cite{ref23} & 29.23 & 0.9160\\
		SRN \cite{ref24} & 30.26 & 0.9342\\
		PSS-NSC \cite{ref25} & 30.92 & 0.9421\\
		DMPHN \cite{ref26} & 31.20 & 0.9451\\
		MT-RNN \cite{ref27} & 31.15 & 0.9450\\
		MPR-Net \cite{ref28} & 32.66 & 0.9589\\
		XYDeblur \cite{ref30} & 30.97 & 0.9501\\
		NAFNet \cite{ref113} & \color{blue}33.69 & \color{blue}0.9668\\
		MSFS-Net \cite{ref114} & 32.73 & 0.9592 \\
		MCMS & \color{red}{33.87} & \color{red}{0.9671}\\
		\hline
	\end{tabular}
\end{table}

Therefore, it can be concluded that the MCMS developed in this study both produces the best outcomes when compared to other approaches after combining the qualitative and quantitative comparison results.

\subsubsection{Comparison Experiments on The RealBlur Dataset}
Next, relevant qualitative and quantitative experiments were conducted on the RealBlur test dataset with the aim of comprehensively evaluating the comprehensive performance of MCMS. Specific experimental results are presented below.

\textbf{Qualitative results.} Fig. 7 shows the test results on the RealBlur test dataset. Observing the processing results of each model in the face region in Fig. 7, it can be found that DeepDeblur and PSS-NSC do improve the clarity compared to the original motion blurred image, but there is still a slight motion blur problem. The processing results of SRN, DMPHN, XYDeblur, MT-RNN, MPR-Net, NAFNet and MSFS-Net, on the other hand, are better than DeepDeblur and PSS-NSC in terms of visualization, but the same slight motion blurring defect exists. Further comparing the performance of each model in the face region, it can be clearly seen that the results of the MCMS developed in this paper exhibit the highest clarity and effectively eliminate the motion blur problem.

In the license plate region of Fig. 7, the differences in the results of the models can be clearly observed. DeepDeblur, SRN, PSS-NSC, MPR-Net, XYDeblur, DMPHN, MT-RNN, and MSFS-Net achieved some deblurring effect, but they still have serious artifacts and motion blurring problems in their processing results. In contrast, the results of NAFNet are improved in terms of visual effects and are better than the results processed by the above algorithms. However, the results of these models still have slight motion blur problems. In comparison, the MCMS model developed in this paper demonstrates significant advantages in terms of clarity and content integrity, and its processing results are more favorable.

\textbf{Quantitative results.} Table 2 demonstrates the quantitative comparison results of different algorithms, which provides a clear data basis for judging the performance of each model. As can be seen from Table 2, the optimal scores in both PSNR and SSIM metrics are obtained by the MCMS model designed in this paper. Compared with the motion deblurring model MSFS-Net, the model in this paper was improved in the PSNR metric and slightly reduced in the SSIM metric.

As mentioned above, the results of the combined qualitative and quantitative evaluation can confirm that the MCMS exhibits the best results through the experiments performed on the RealBlur dataset.

\begin{table}[h]
	\caption{Quantitative results of different methods on the RealBlur dataset. Red values indicate the best results, blue values represent sub-optimal performance\label{tab:table2}}
	\centering
	\begin{tabular}{c c c}
		\hline
		Methods & PSNR ($\uparrow$)& SSIM ($\uparrow$)\\
		\hline
		DeepDeblur \cite{ref23} & 27.87 & 0.8270\\
		SRN \cite{ref24} & 28.56 & 0.8671\\
		PSS-NSC \cite{ref25} & 26.52 & 0.8570\\
		DMPHN \cite{ref26} & 28.42 & 0.8602\\
		MT-RNN \cite{ref27} & 28.44 & 0.8620\\
		MPR-Net \cite{ref28} & 28.70 & 0.8731\\
		XYDeblur \cite{ref30} & 26.85 & 0.8593\\
		NAFNet \cite{ref113} & 28.32 & 0.8570\\
		MSFS-Net \cite{ref114} & \color{blue}28.97 & \color{red}0.9080 \\
		MCMS & \color{red}{29.13} & \color{blue}{0.8936}\\
		\hline
	\end{tabular}
\end{table}

\subsubsection{Comparison Experiments on The REBI Dataset}
Since the RWBI dataset does not contain the corresponding ground truth, the corresponding qualitative experiment is the only feasible way to evaluate it in this part of the comparison experiment.

\textbf{Qualitative results.} Fig. 8 demonstrates the test results on the RWBI dataset. Observing the processing results of each model in the English word region in Fig. 8, it can be found that DeepDeblur and MSFS-Net show some motion deblurring effect when processing large font letters, while no significant improvement is seen for small font letters.The motion deblurring effect of SRN, PSS-NSC, DMPHN, MT-RNN, MPR-Net, NAFNet and XYDeblur do not have significant motion deblurring effects. It can be clearly seen through careful observation that the MCMS model developed in this paper exhibits the highest clarity and achieves the optimal processing results.

In the white-framed glass wall region of Fig. 8, the differences in the results of the models are obvious.The processing results of DeepDeblur, SRN and PSS-NSC show more obvious distorted regions, which significantly affect the image integrity. Although the processing results of DMPHN, MT-RNN and XYDeblur have serious detail loss. the results of MPR-Net, NAFNet and MSFS-Net are relatively clearer, but the motion deblurring effect in the detail region is not outstanding. In contrast, the MCMS model designed in this paper performs better in terms of clarity and content integrity, and its processing results are superior.

\begin{figure*}[b]
	\centering
	\includegraphics[width=5in]{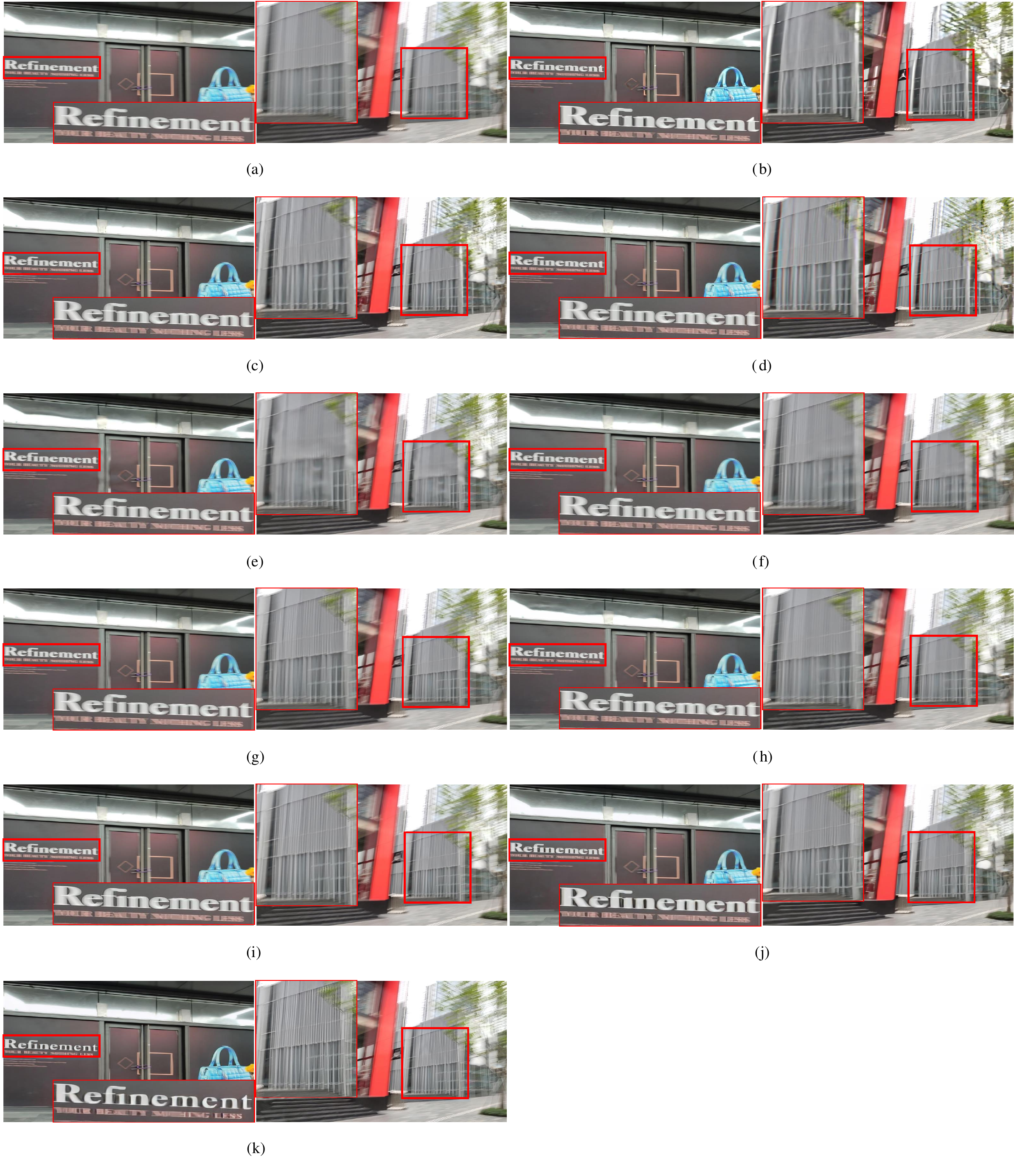}
	\caption{Qualitative results of different plans on the RWBI dataset. (a) Blurry images. (b) DeepDeblur \cite{ref23}. (c) SRN \cite{ref24}. (d) PSS-NSC \cite{ref25}. (e) DMPHN \cite{ref26}. (f) MT-RNN \cite{ref27}. (g) MPR-Net \cite{ref28}. (h) XYDeblur \cite{ref30}. (i) NAFNet \cite{ref113}. (j) MSFS-Net \cite{ref114}. (k) MCMS.}
	\label{fig8}
\end{figure*}

\subsection{Ablation Analysis}
In this paper, ablation experiments are implemented on the GoPro test dataset to validate the performance of each module. The grouped feature fusion module significantly enhances the feature extraction capability of MCMS by accurately extracting feature information in channel and spatial dimensions. Meanwhile, the MSSA module further enhances the feature representation capability of MCMS by skillfully combining the anisotropy and multi-scale information of images. These two modules play a key role in enhancing the performance of MCMS.

The ablation study is shown below:

\begin{itemize}
	\item -w Grouped feature fusion, MCMS contains only the grouped feature fusion module;
	
	\item -w MSSA, MCMS contains only MSSA.
\end{itemize}

\textbf{Qualitative results.} Fig. 9 demonstrates the results of motion deblurring. By comparing the original motion blurred images, it can be found that although both -w grouped feature fusion and -w MSSA have achieved some deblurring effect, there are still obvious distorted pixels. In contrast, the MCMS model applying all modules performs well in motion deblurring and significantly improves the clarity and overall quality of the image.

\begin{figure}[h]
	\centering
	\includegraphics[width=3.4in]{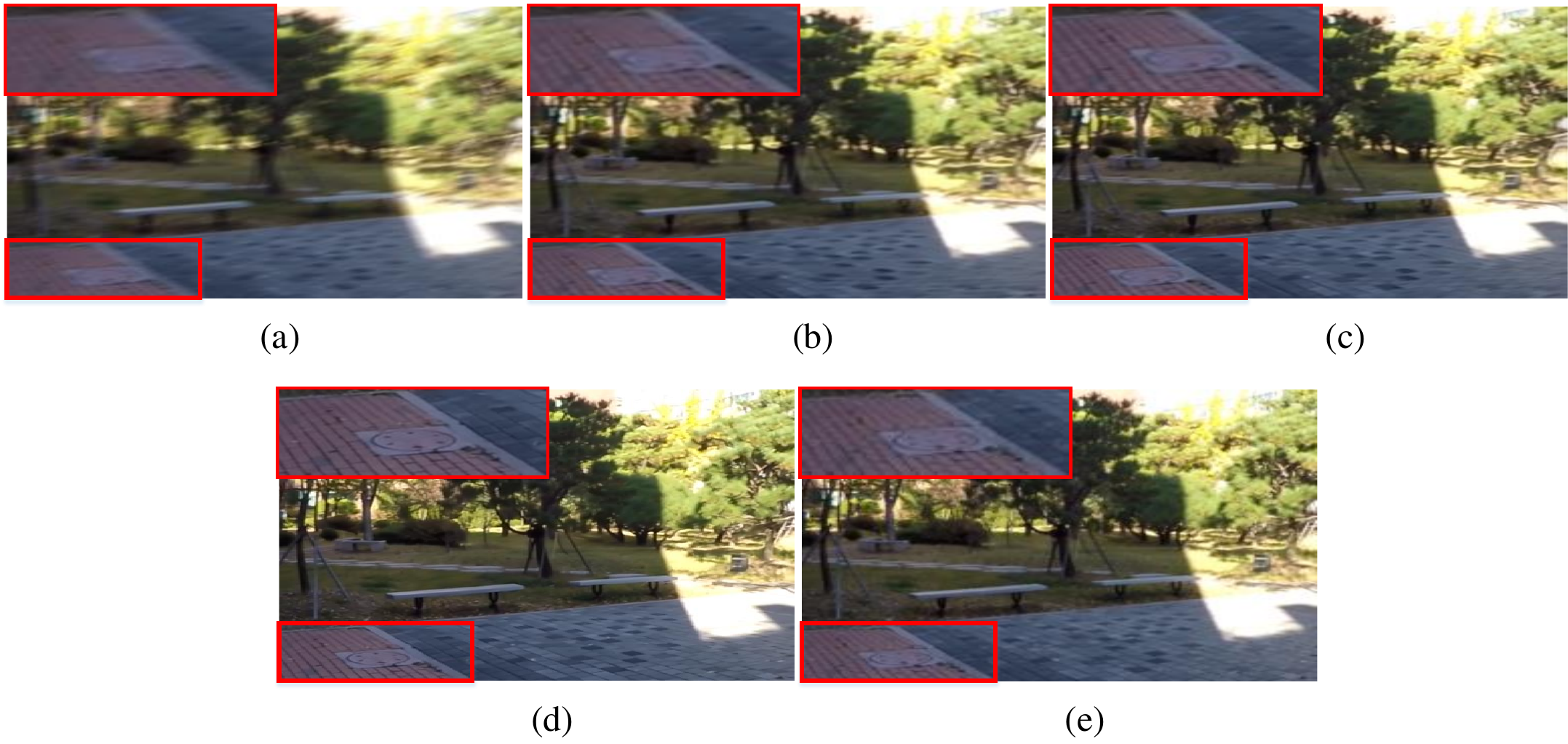}
	\caption{Ablation comparison results. (a) Blurry images. (b)  Results of -w Grouped feature fusion. (c) Results of -w MSSA. (d) Includes results from all modules. (e) Ground-truth}
	\label{fig9}
\end{figure}

\textbf{Quantitative results.}

Table 3 details the quantitative results of the defuzzification study, which provides intuitive data to support the model performance. Each of the three experiments carried out in this paper delves into the role of different modules in terms of motion deblurring effects. In the first experiment, the model uses only the grouped feature fusion module and obtains a PSNR value of 33.69 and an SSIM value of 0.9630 on the GoPro test dataset.In the second experiment, the model uses only the MSSA module and obtains a PSNR value of 33.75 and an SSIM value of 0.9648 on the same dataset.As can be clearly observed from Table 3, although the both the first and second experiments have been successful, the highest PSNR and SSIM values of DNMCMS are achieved when both modules are used simultaneously. This fully proves the importance of the combination of the grouped feature fusion module and the MSSA module for improving the model motion deblurring performance.

\begin{table}[!t]
	\setlength\tabcolsep{4pt} 
	\caption{Quantitative comparison outcomes of diverse components on the GoPro dataset.}
	\centering
	\label{tab:UGAN and UIEBD}
	\begin{tabular}{cc|cc}
		\hline
		\multicolumn{2}{c}{Distillation component}&\multicolumn{2}{c}{Assessment of indicators}\\
		Grouped feature fusion&MSSA&PSNR&SSIM\\
		\hline
		$\checkmark$&$\times$ &33.69&0.9630\\
		$\times$&$\checkmark$&33.75&0.9648\\
		$\checkmark$&$\checkmark$&\color{red}33.87&\color{red}0.9671\\
		\hline
	\end{tabular}
\end{table}

\section{Conclusion}
In this paper, a three-stage encoder-decoder model is designed based on the unique characteristics of high-frequency and low-frequency components to deal with the motion blur problem of images more effectively. The model is able to extract the edge information of the high-frequency component and the structural information of the low-frequency component, so as to improve the quality of the image in a specific stage. Through this three-stage design, this paper is able to better utilize the information of different frequency components in the image, thus improving the motion deblurring capability.

In addition, this paper develops a grouped feature fusion technique that aims to comprehensively fuse various types of features. This technique can effectively integrate feature information from different stages, enabling the model to understand and process the image more comprehensively. Meanwhile, an MSSA module is designed which significantly enhances the feature representation capability of the deep model. This module empowers the model to adaptively focus on different parts of the image at different scales, which in turn captures the details and structural information in the image more accurately.

\section*{Acknowledgments}
There is no financial support for this work. The authors declare that no conflicts of interest exist.


%

\end{document}